\documentclass[runningheads]{llncs}
\usepackage[T1]{fontenc}
\usepackage{graphicx}
\usepackage{amsmath,amssymb}
\usepackage{commath}
\usepackage{physics}
\usepackage[pagebackref,breaklinks,colorlinks]{hyperref}
\usepackage{makecell}
\usepackage{color}
\usepackage{breakcites}
\usepackage{placeins}

\begin{document}
\title{Learning Differential Invariants of Planar Curves}

\author{Roy Velich \and
Ron Kimmel}
\authorrunning{R. Velich et al.}
%
\institute{Technion - Israel Institute of Technology \\
\email{royve@campus.technion.ac.il}\\
\email{ron@cs.technion.ac.il}}



\maketitle              
\begin{abstract}
We propose a learning paradigm for the numerical approximation of differential invariants of planar curves.   
Deep neural-networks' (DNNs) universal approximation properties are utilized to estimate geometric measures. 
The proposed framework is shown to be a preferable alternative to axiomatic constructions.
Specifically, we show that DNNs can learn to overcome instabilities and sampling artifacts and produce consistent signatures for curves subject to a given group of transformations in the plane.
We compare the proposed schemes to alternative state-of-the-art axiomatic constructions of differential invariants. We evaluate our models qualitatively and quantitatively and propose a benchmark dataset to evaluate approximation models of differential invariants of planar curves.

\end{abstract}
\section{Introduction}
    \label{sec:intro}
    \textit{Differential Invariants and Signature Curves.}
        According to an important theorem by É. Cartan \cite{cartan1935}, two curves are related by a group transformation $g \in G$, i.e., equivalent curves with respect to $G$, if and only if their signature curves, with respect to the transformation group $G$, are identical. 
        This observation allows one to develop analytical tools to measure the equivalence of two planar curves, which has practical applications in various fields and tasks, such as computer vision, shape analysis, geometry processing, object detection, and more. 
        Planar curves are often extracted from images as boundaries of objects, or level-sets of gray-scale images. 
        The boundary of an object encodes vital information which can be further exploited by various computer vision and image analysis tools. 
        A common computer vision task requires one to find common characteristics between two objects in two different image frames. 
        This task can be naturally approached by representing image objects as signature curves of their boundaries. Since signature curves are proven to provide a full solution for the equivalence problem of planar curves \cite[p.~183, Theorem 8.53]{olver_1999}, they can be further analyzed for finding correspondence between image objects.
        Signature curves are parametrized by differential invariants. 
        Therefore, in order to generate the signature curve of a given planar curve, one has to evaluate the required differential invariants at each point. 
        In practice, planar curves are digitally represented as a discrete set of points, which implies that the computation of differential invariant quantities, such as the curvature at a point, can only be numerically approximated using finite differences techniques.
        Since many important and interesting differential invariants are expressed as functions of high-order derivatives, their approximations are prone to inherent numerical instabilities, and high sensitivity to sampling noise.
        For example, the equiaffine curvature, which is a differential invariant of the special-affine transformation group $\mathrm{SA}\left(2\right)$, is a fourth-order differential invariant, and its direct approximation using an axiomatic approach is practically infeasible for discrete curves, especially around inflection points. 
        
    \textit{Contribution.}
        Inspired by the numerical approximation efforts of Calabi et al. \cite{Calabi98differentialand}, and the learning approaches introduced in \cite{pai2017learning,Lichtenstein2019}, we present a complete learning environment for approximating differential invariants of planar curves with respect to various transformation groups, including the affine group, for which, very few attempts to approximate their differential invariants in the discrete setting have been made. Moreover, we introduce a shape-matching benchmark for evaluating differential invariants approximation models.

\section{Related Work}
    \label{sec:related_work}
    \textit{Axiomatic Approximation of Differential Invariants.}
        The axiomatic approach for approximating the differential invariants of discrete planar curves, as presented by Calabi et al. in \cite{Calabi98differentialand}, is based on the joint invariants between points of a discrete planar curve. 
        As suggested by Calabi et al., given a transformation group $G$ and a planar curve $\mathcal{C}$, one can approximate the fundamental differential invariant w.r.t. $G$ at a point $\vb{x} \in \mathcal{C}$, which is denoted by $\kappa\left(\vb{x}\right)$ and is known as the $G$-\textit{invariant curvature} at $\vb{x}$, by first interpolating a set $\mathcal{P}_{\vb{x}} \subset \mathcal{C}$ of points from a small neighborhood around $\vb{x}$ with an auxiliary curve $\mathcal{C}_{\vb{x}}$, on which the differential invariant is constant by definition. Then, the constant differential invariant $\widetilde{\kappa}\left(\vb{x}\right)$ of $\mathcal{C}_{\vb{x}}$ is evaluated using the joint invariants of the points in $\mathcal{P}_{\vb{x}}$. Finally, $\widetilde{\kappa}\left(\vb{x}\right)$ is used as an approximation to $\kappa\left(\vb{x}\right)$.
        Since the approximation is calculated by joint invariants of $G$, its evaluation is unaffected by the action of a group transformation $g \in G$, and therefore, as the size of the mesh of points $\mathcal{P}_{\vb{x}}$ tends to zero, the approximation of the differential invariant at $\vb{x}$ converges to the continuous value. 
        For example, in the Euclidean case, the fundamental joint invariant of the groups $\mathrm{E}\left(2\right)$ and $\mathrm{SE}\left(2\right)$ is the Euclidean distance between a pair of points. By applying the approach suggested by Calabi et al., the Euclidean curvature $\kappa\left(\vb{x}\right)$ can be approximated by first interpolating a circle through $\vb{x_{i-1}}$, $\vb{x_i}$, and $\vb{x_{i+1}}$, and then exploit the Euclidean distances between those three points to calculate the constant curvature $\widetilde{\kappa}\left(\vb{x}\right)$ of the interpolated circle, using Heron's formula.
        The second differential invariant at a curve point $\vb{x}$, is defined as the derivative of the $G$-invariant curvature $\kappa$ with respect to the $G$-\textit{invariant arc-length element} $s$, and is denoted by $\kappa_s\left(\vb{x}\right)$. According to Calabi et al., $\kappa_s\left(\vb{x}\right)$ is approximated using finite differences, by calculating the ratio of the difference between $\widetilde{\kappa}\left(\vb{x}_{i-1}\right)$ and $\widetilde{\kappa}\left(\vb{x}_{i+1}\right)$ with respect to the distance between $\vb{x_{i-1}}$ and $\vb{x_{i+1}}$. 
        In their paper, Calabi et al. further explain how to approximate  differential invariants of planar curves with respect to the equiaffine group $\mathrm{SA}\left(2\right)$.
        In this case, the fundamental joint invariant is the triangle area defined by a triplet of points, and the auxiliary curve is a conic section, which possesses a constant equiaffine curvature, and has to be interpolated through five curve points in the neighborhood of $\vb{x}$. 
        The main limitation of their method, in the equiaffine case, is that it is practically applicable only to convex curves, since the equiaffine curvature is not defined at inflection points.
        
    \textit{Learning-Based Approximation.}
        A more recent approach was introduced in \cite{pai2017learning}. 
        There, Pai et al. took a self-supervised learning approach and used a Siamese convolutional neural network to learn the euclidean curvature from a dataset of discrete planar curves. 
        However, the scope of their work was lacking a module to approximate the Euclidean arc-length at each point and/or the derivative of the curvature w.r.t. arc-length, and therefore, a signature curve could not be generated. This means that their approach was not useful for evaluating the equivalence between planar shapes. Moreover, they have not extended their work to learning differential invariants of less restrictive transformation groups, such as the equiaffine and affine groups. 
        In \cite{Lichtenstein2019}, Lichtenstein et al. introduced a deep learning approach to numerically approximate the
        solution to the Eikonal equation.
        They proposed to replace axiomatic local numerical solvers with a trained neural network that provides highly accurate estimates of local distances for a variety of different geometries and sampling conditions.
        %

\section{Mathematical Framework}

    We begin with the following theorem, as stated by \cite[p.~176, Theorem 8.47]{olver_1999},
    \begin{theorem}[The Fundamental Theorem of Differential Invariants]
    A transformation group $G$ acting on $\mathbb{R}^2$ admits the following properties. A unique differential invariant $\kappa$ (up to a function of it), which is known as the $G$-invariant curvature. A unique $G$-invariant one-from $ds = p\left(x\right)dx$ (up to a constant multiple), which is known as the $G$-invariant arc-length element. Any other differential invariant of $G$ is a function $I: \left(\kappa, \kappa_s, \kappa_{ss}, \hdots \right) \rightarrow \mathbb{R}$ of the $G$-invariant curvature $\kappa$ and its derivatives with respect to the $G$-invariant arc length $s$.
    \end{theorem}
    \begin{definition}[Equivalent Planar Curves]
    Let $G$ be a transformation group acting on $\mathbb{R}^2$. Two planar curves $\mathcal{C}$ and $\hat{\mathcal{C}}$ are \textit{equivalent} with respect to $G$ if there exists $g \in G$ such that $\hat{\mathcal{C}} = g \cdot \mathcal{C}$.
    \end{definition}
    \begin{definition}[Signature Curve]
    The $G$-invariant signature curve associated with a parametrized planar curve $\mathcal{C} = \left\{\left(x\left(t\right), y\left(t\right)\right)\right\}$, is a curve $S \subset \mathbb{R}^2$ parametrized by the $G$-invariant curvature of $\mathcal{C}$ at a point and its derivative with respect to the $G$-invariant arc-length, and is given by $S = \left\{\left(\kappa\left(t\right), \kappa_s\left(t\right)\right)\right\}$.
    \end{definition}
    Signature curves can be used to solve the equivalence problem of planar curves for a general transformation group $G$, as stated by the next theorem, given in \cite[p.~183, Theorem 8.53]{olver_1999},
    \begin{theorem}[Equivalence of Planar Curves]
    Two non-singular planar curves $\mathcal{C}$ and $\hat{\mathcal{C}}$ are equivalent w.r.t. a transformation group $G$ if and only if their signature curves $S$ and $\hat{S}$ are equal.
    \end{theorem}
    For more information about differential invariants and signature curves, see \cite{OlverSapiroTannenbaum1994,olver_1995,olver_1999,BrucksteinHoltNetravaliRichardson1992,BrucksteinKatzirLindenbaumPorat1992,BrucksteinNetravali1995,kimmel1996,KimmelZhangBronsteinABronsteinM2011, BRUCKSTEIN1998181,CattLionsMorelColl1992,BrucksteinHoltNetravaliRichardson1992}.

\section{Method}
    We introduce a simple neural-net architecture, along with an appropriate training scheme and loss function, for producing numerically stable approximations of the two fundamental differential invariants $\kappa$ and $\kappa_s$, with respect to a group transformation $G$. 
    Both quantities can be combined together to evaluate and plot the $G$-invariant signature curve of a planar curve. 
    Given a planar curve $\mathcal{C}$ and a point $\vb{p} \in \mathcal{C}$ on the curve, the neural network receives as an input a discrete sample of the local neighborhood of $\vb{p}$, and outputs two scalars, of which we interpret as $\kappa$ and $\kappa_s$ at $\vb{p}$.
    \subsection{Learning Differential Invariants}
        There are two fundamental ideas that govern our training approach for learning a differential invariant w.r.t. a transformation group $G$.
        \label{sec:learning_curvature}
        
        \textit{Group and Reparametrization Invariance.}
            The neural network should learn a representation that is both invariant to the action of a group transformation $g \in G$, and to a reparametrization of the input curve. That way, the network will learn a truly invariant representation, which is not biased towards any specific discrete sampling scheme, and that encapsulates the local geometric properties of the curve.
            Given a discrete planar curve $\mathcal{C}$, we denote by $\mathcal{C}\left(\mathcal{D}\right)$ its down-sampled version, sampled non-uniformly with $N$ points according to a random non-uniform probability mass function $\mathcal{D}: \mathcal{C} \rightarrow \left[0,1\right]$.
            Given two different non-uniform probability mass functions $\mathcal{D}_1$ and $\mathcal{D}_2$,
            one can refer to the down-sampled curves $\mathcal{C}\left(\mathcal{D}_1\right)$ and $\mathcal{C}\left(\mathcal{D}_2\right)$, as two different reparametrizations of $\mathcal{C}$.
            This motivates us to require that if a point $\vb{p} \in \mathcal{C}$ was drawn both in $\mathcal{C}\left(\mathcal{D}_1\right)$ and $\mathcal{C}\left(\mathcal{D}_2\right)$, then, a valid prediction model should output the same curvature value for $\vb{p}$, whether it was fed with either a neighborhood of $g_1 \cdot \vb{p}$ or a neighborhood $g_2 \cdot \vb{p}$ as an input, for any two group transformations $g_1, g_2 \in G$.
            
\begin{figure*}[t!]
            \centering
            \includegraphics[width=\textwidth]{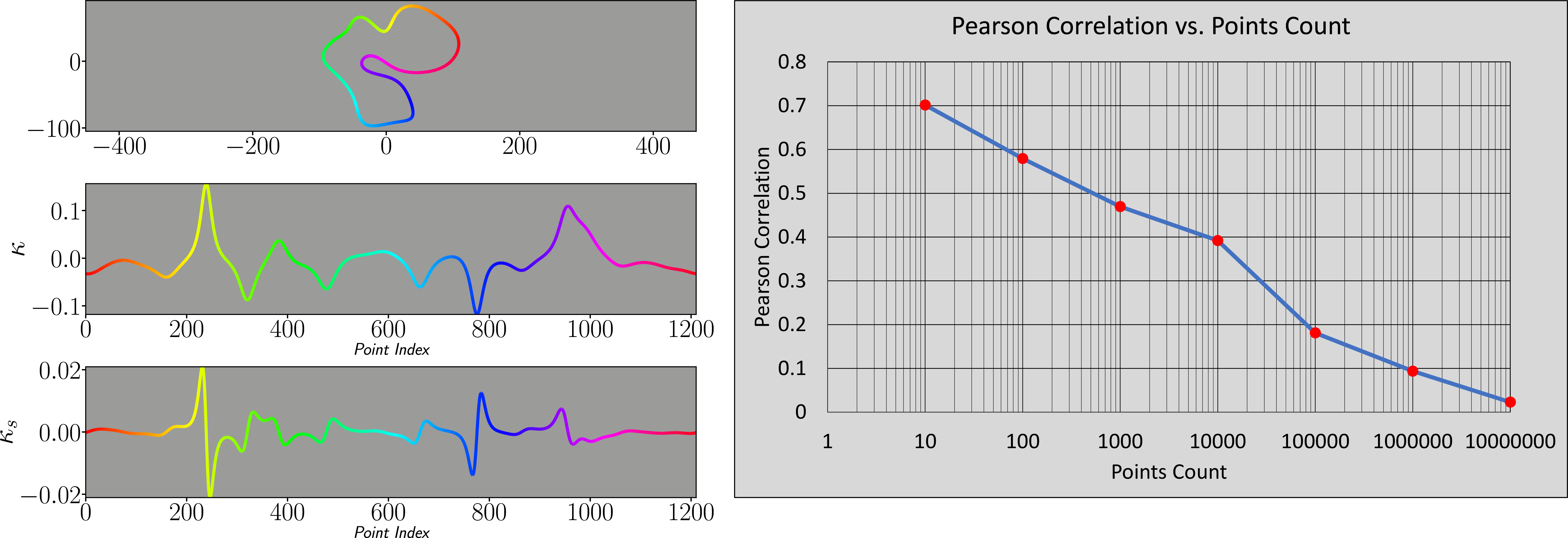}
            \caption{\textbf{Left}: Plot of a smooth curve and its Euclidean $\kappa$ and $\kappa_s$ evaluated at each point. \textbf{Right}: A semi-log plot of the absolute value of the Pearson correlation coefficient of $X_{\kappa}$ and $X_{\kappa_s}$ as a function of points count.}
            \label{fig:pearson}
        \end{figure*}

        \textit{Orthogonality of Differential Invariants.}
            We expect $\kappa$ and $\kappa_s$ to be orthogonal quantities, since the latter is the derivative of the former w.r.t. to arc-length. So, for example, consider a collection of $M$ points, such that each was sampled from an arbitrary planar curve. Moreover, let $\kappa^i$ and $\kappa_s^i$ denote the curvature and its derivative w.r.t. arc-length evaluated at the $i^{th}$ point. Then, if we consider $\left\{\kappa^i\right\}_{i=1}^M$ and $\left\{\kappa_s^i\right\}_{i=1}^M$ as $M$ observations of random variables $X_{\kappa}$ and $X_{\kappa_s}$, respectively, we expect the absolute value of the Pearson correlation coefficient $\abs{\rho\left(X_{\kappa}, X_{\kappa_s}\right)}$ to approach zero as $M \to \infty$. We empirically verify this assumption in the Euclidean case, by calculating the Pearson correlation coefficient of $X_{\kappa}$ and $X_{\kappa_s}$ on a collection of randomly sampled points, taken from a large collection of smooth curves. See Figure \ref{fig:pearson}.

    \subsection{Training Scheme}
        \label{sec:learning_curvature_training_scheme}
        Given a dataset of $n$ discrete planar curves $\left\{\mathcal{C}_i\right\}_{i=1}^n$, 
        our training scheme generates a training batch on the fly, where each batch is made of a set of training tuplets. 
        A training tuplet $\mathcal{T} = \left(\mathcal{E}_a, \mathcal{E}_p, \mathcal{E}_{n_1}, \hdots, \mathcal{E}_{n_{m}}\right)$, is made up of a collection of sampled neighborhoods, which are referred to as the \textit{anchor} ($\mathcal{E}_a$), \textit{positive} ($\mathcal{E}_p$) and \textit{negative} ($\left\{\mathcal{E}_{n_i}\right\}_{i=1}^{m}$) examples. 
        The tuplet is generated by first drawing a random curve $\mathcal{C}$ and a random point $\vb{p} \in \mathcal{C}$ on it. 
        Given two random group transformations $g_a, g_p \in G$ and two random non-uniform probability mass functions $\mathcal{D}_a$ and $\mathcal{D}_p$, both the anchor and positive examples, $\mathcal{E}_a$ and $\mathcal{E}_p$, are generated by selecting $2N$ adjacent points to $\vb{p}$ ($N$ consecutive points that immediately precede $\vb{p}$, and additional $N$ consecutive points that immediately succeed $\vb{p}$) from the transformed and down-sampled versions of $\mathcal{C}$ given by $g_a \cdot \mathcal{C}_a\left(\mathcal{D}_a\right)$ and $g_p \cdot \mathcal{C}_p\left(\mathcal{D}_p\right)$, respectively. 
        In other words, both $\mathcal{E}_a$ and $\mathcal{E}_p$ represent a local neighborhood of $\vb{p}$ sampled under different reparametrizations, which were transformed by group transformations taken from the same transformation group $G$. 
        
        The $i^{\mathrm{th}}$ negative example $\mathcal{E}_{n_i}$ is generated in a similar manner. We draw another random curve $\mathcal{C}_i$, and take a random point $\vb{p}_i \in \mathcal{C}_i$ on it. Then, we select again additional $2N$ adjacent points to $\vb{p}_i$ from a transformed and down-sampled version of $\mathcal{C}_i$ given by $g_{n_i} \cdot \mathcal{C}_i\left(\mathcal{D}_i\right)$, where $g_{n_i} \in G$ is a random group transformation, and $\mathcal{D}_i$ is a random probability mass function. 
        
        Given a batch of $K$ training tuplets $\left\{\mathcal{T}_i\right\}_{i=1}^K$, we train a multi-head Siamese fully-connected neural network by minimizing the loss given by,
        \begin{eqnarray}
            \label{eq:loss}
            \mathcal{L} &=& \mathcal{L}_I + \mathcal{L}_O
        \end{eqnarray}
        Such that $\mathcal{L}_I$ and $\mathcal{L}_O$ are given by,
            \begin{align}
                \label{eq:invariance_loss}
                &\mathcal{L}_I = \frac{1}{K} \sum_{i=1}^K \mathcal{L}_{\text{tuplet}}^i \\
                \label{eq:tuplet_loss}
                &\mathcal{L}_{\text{tuplet}}^i = \log\left(1+\sum_{j=1}^{m} \exp\left(\lVert \textbf{x}^i_a - \textbf{x}^i_p \rVert - \lVert \textbf{x}^i_a - \textbf{x}^i_{n_j} \rVert \right) \right) \\
                \label{eq:orthogonality_loss}
                &\mathcal{L}_{O} = \frac{1}{m+2} \left(\abs{\rho\left(\left\{\textbf{x}_a^i\right\}_{i=1}^K\right)} + \abs{\rho\left(\left\{\textbf{x}_p^i\right\}_{i=1}^K\right)} + \sum_{j=1}^m \abs{\rho\left(\left\{\textbf{x}_{n_j}^i\right\}_{i=1}^K\right)}\right) \\
                \label{eq:pearson_corr}
                &\rho\left(\left\{\vb{x}^i\right\}_{i=1}^K\right) = \frac{\sum_{i=1}^K \left(\vb{x}^i\left[0\right] - \bar{\vb{x}}\left[0\right]\right)\left(\vb{x}^i\left[1\right] - \bar{\vb{x}}\left[1\right]\right)}{\sqrt{\sum_{i=1}^K \left(\vb{x}^i\left[0\right] - \bar{\vb{x}}\left[0\right]\right)^2 \sum_{i=1}^K \left(\vb{x}^i\left[1\right] - \bar{\vb{x}}\left[1\right]\right)^2}}
            \end{align}

\begin{figure*}[t!]
            \centering
            \includegraphics[width=\textwidth]{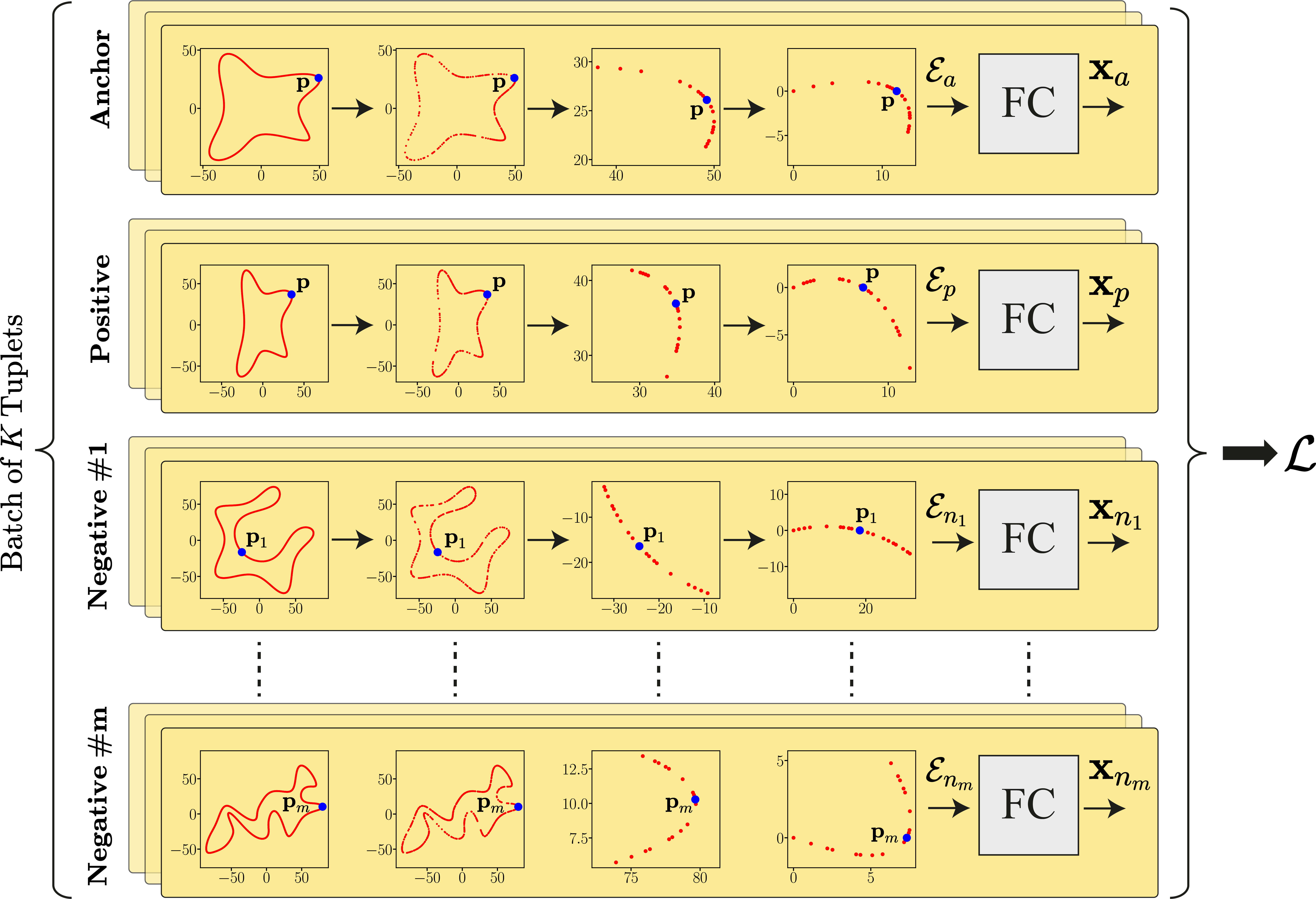}
            \caption{A demonstration of the training scheme for a given input batch of training tuplets. \textbf{Left-to-right}: A random curve and a random point on it are drawn. The curve is down-sampled non-uniformly. A local neighborhood around the point is extracted. The extracted point neighborhood is transformed into a canonical placement, as described above. The canonical point neighborhood is fed into the neural-network, which outputs two scalars that we interpret as $\kappa$ and $\kappa_s$. \textbf{Top-to-bottom}: The anchor and positive training examples, as well as the $m$ negative training examples. Together, they form a training tuplet.}
            \label{fig:curvature_tuplet}
        \end{figure*}

        
        Where $\textbf{x}_a^i$, $\textbf{x}_p^i, \textbf{x}_{n_1}^i, \hdots, \textbf{x}_{n_{m}}^i \in \mathbb{R}^2$ are the neural-network's outputs for the $i^{th}$ input tuplet given by $\mathcal{T}_i = \left(\mathcal{E}_a^i, \mathcal{E}_p^i, \mathcal{E}_{n_1}^i, \hdots, \mathcal{E}_{n_{m}}^i\right)$, and where $\vb{x}\left[0\right]$ and $\vb{x}\left[1\right]$ are the first and second components of an output vector, respectively. Moreover, $\bar{\vb{x}}\left[0\right]$ and $\bar{\vb{x}}\left[1\right]$ are the mean values of the first and second components of a set of output vectors, respectively. Equation \ref{eq:tuplet_loss} is known as the \textit{tuplet loss} \cite{9010708} which is minimized when the distance between $\textbf{x}_a^i$ and $\textbf{x}_p^i$ is minimized, and the distance between $\textbf{x}_a^i$ and $\textbf{x}_{n_j}^i$ is maximized, for $j=1, \hdots, m$. Equation \ref{eq:invariance_loss} is named the \textit{invariance loss} and it is equal to the mean of the tuplet loss over all the training tuplets in the training batch. Equation \ref{eq:orthogonality_loss} is named the \textit{orthogonality loss}, and it is minimized when the absolute value of the Pearson correlation coefficient is minimized w.r.t. the anchor, positive, and negative examples' outputs, over the whole batch.
        See Figure \ref{fig:curvature_tuplet} for an elaborate visual demonstration of the training scheme for a batch of training tuplets.
        Note, that we eliminate any translation and rotational ambiguity by transforming each input example into a canonical placement before it is fed into the neural-network. 
        The transformation is done by translating each example such that its first point is located at the origin and then, rotating it such that the vector that starts at the example's first point and ends at the middle point is aligned with the positive x-axis.
    \section{Experiments and Results}
        \subsection{Datasets and Training}
            We create three datasets of planar curves, by extracting level-curves from arbitrary images scrapped from the internet. The first two datasets are used for training and validation, and the third dataset is used for qualitative evaluation. Each dataset consists of tens of thousands of curves. We train a multi-head siamese neural-network based on a simple fully-connected architecture. The architecture is based on blocks of 3 fully-connected layers of the same size. The layers at the first block are of size 128, and the size of the layers in each subsequent block is divided by two w.r.t. the size of layers in the previous block. We use a batch-norm layer after each linear layer, followed by a sine activation function. The reason for using a periodic activation function is due to the fact that its derivative can be expressed by a phase-shift. We believe that this fact makes it easier for our network to learn high-order derivatives \cite{sitzmann2019siren}.
        \subsection{Qualitative Evaluation}
            We evaluate our differential invariants approximation model qualitatively, w.r.t. the affine group. We focus on interpretability, group invariance, reparametrization invariance, and comparison against the Euclidean axiomatic signature.
            
\begin{figure*}[t!]
                    \centering
                    \includegraphics[width=\textwidth]{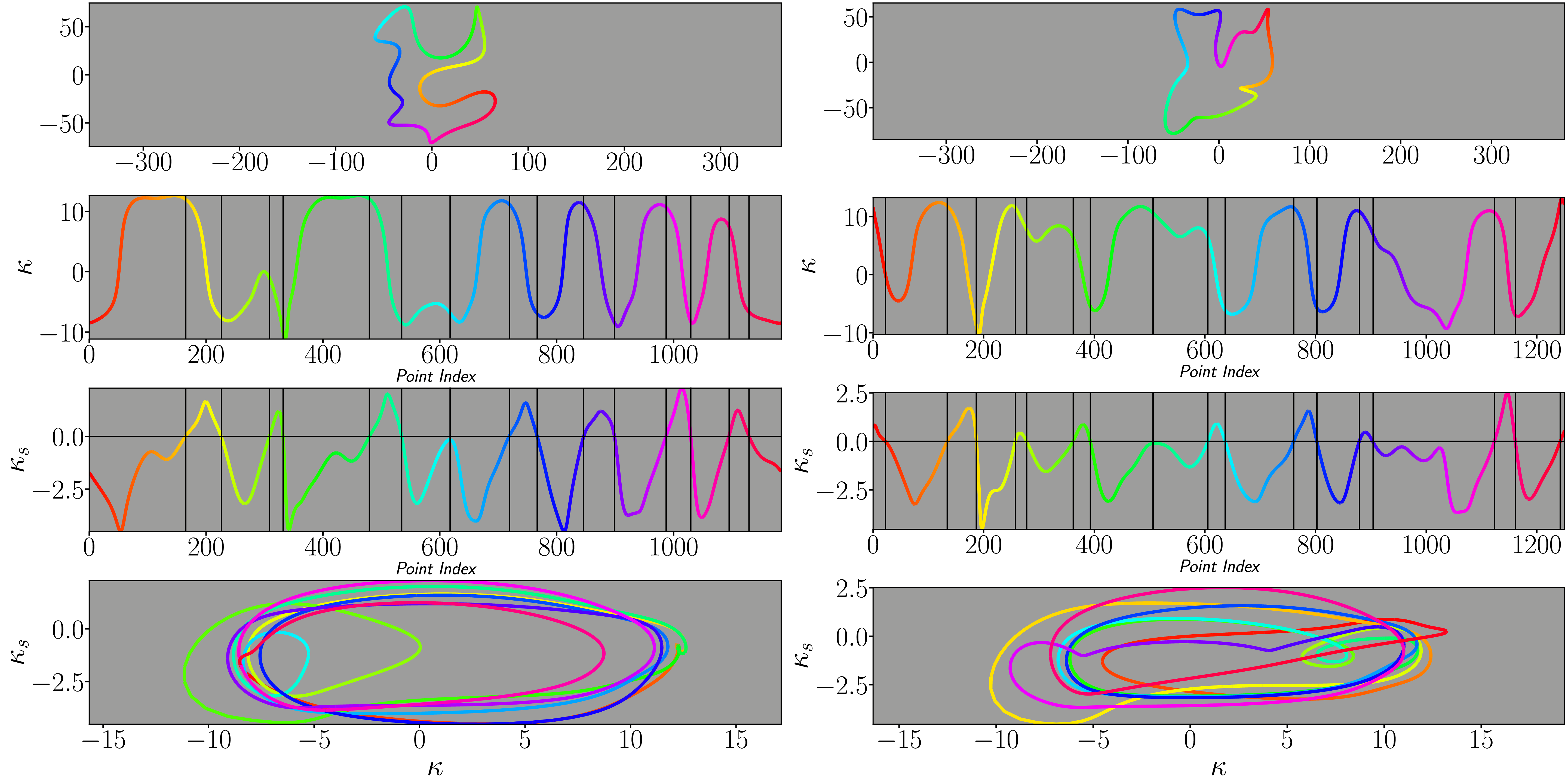}
                    \caption{Two discrete planar curves and the plots of our model's approximation for their affine $\kappa$ and $\kappa_s$ as a function of point-index, and $\kappa_s$ as a function of $\kappa$ (affine signature curve).}
                    \label{fig:interpretability}
                \end{figure*}

            \textit{Interpretability.}
                As shown in Figure \ref{fig:interpretability}, our model's outputs can be interpreted as $\kappa$ and $\kappa_s$. The first output, which we interpret as $\kappa$, approximately reaches its extrema points when the second output, which we interpret as $\kappa_s$, crosses the zero line.
                
\begin{figure*}[t!]
                    \centering
                    \includegraphics[width=\textwidth]{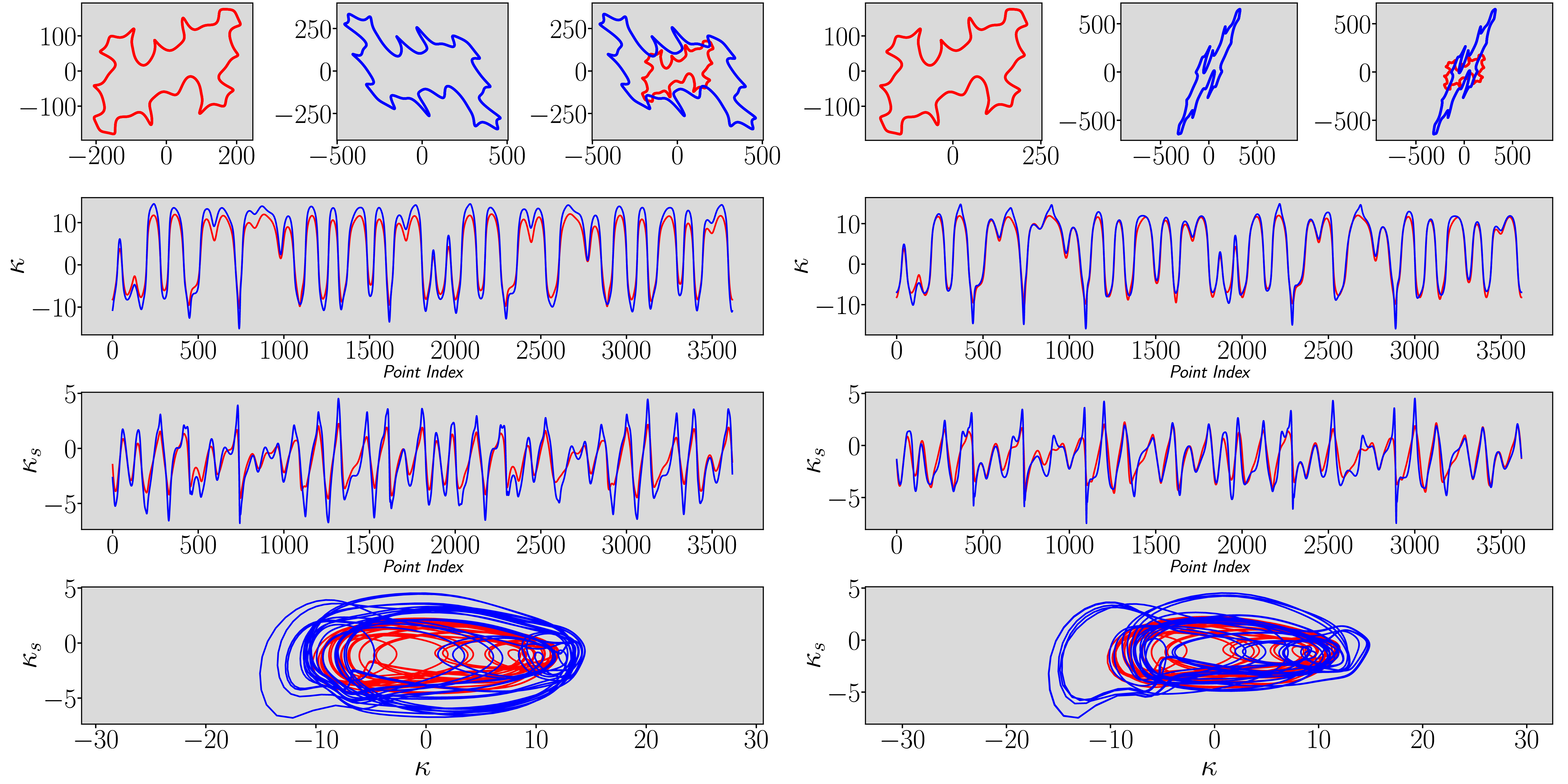}
                    \caption{A comparison of our model's approximation for the affine $\kappa$ and $\kappa_s$ as a function of point-index, of two pairs of equivalent curves. The blue curve was obtained by applying a linear operator on the red one. \textbf{Left}: Linear operator with condition number 2.5 and determinant 3.5. \textbf{Right}: Linear operator with condition number 5 and determinant 2.}
                    \label{fig:group_invariance1}
                \end{figure*}

            \textit{Group Action Invariance.}
                As shown in Figure \ref{fig:group_invariance1}, our approximation model exhibits clear group invariance w.r.t. the affine group. The peaks and valleys of the plots of $\kappa$ and $\kappa_s$ of the two equivalent curves (the red and blue curves) are aligned correctly, and the relative amplitude amplification is mild.
                
\begin{figure*}[t!]
                    \centering
                    \includegraphics[width=\textwidth]{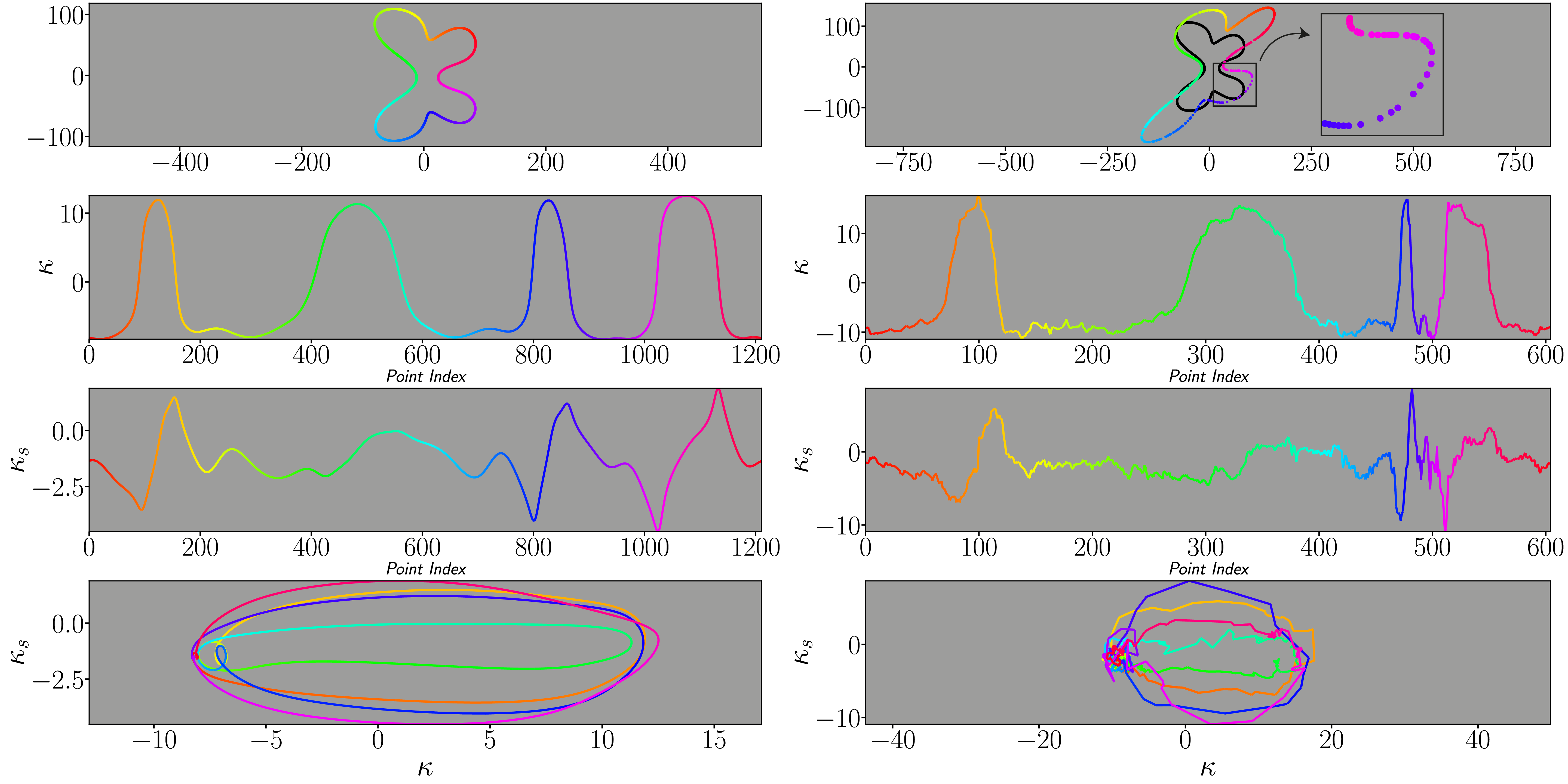}
                    \caption{A demonstration of our model's robustness for non-uniform down-sampling. \textbf{Left}: Our model's approximation for $\kappa$ and $\kappa_s$ of a reference planar curve. \textbf{Right}: Our model's approximation for $\kappa$ and $\kappa_s$ of the reference curve, after it was non-uniformly down-sampled by 50\%, and an arbitrary affine transform was applied to it.}
                    \label{fig:reparam_invariance}
                \end{figure*}

            \textit{Reparametrization Invariance.}
                As shown in Figure \ref{fig:reparam_invariance}, our approximation model is robust to reparametrization and non-uniform down-sampling. The peaks and valleys patterns of $\kappa$ and $\kappa_s$ w.r.t. the reference curve are clearly reproduced in its down-sampled and transformed version (matching colors).
                
\begin{figure*}[t!]
                    \centering
                    \includegraphics[width=\textwidth]{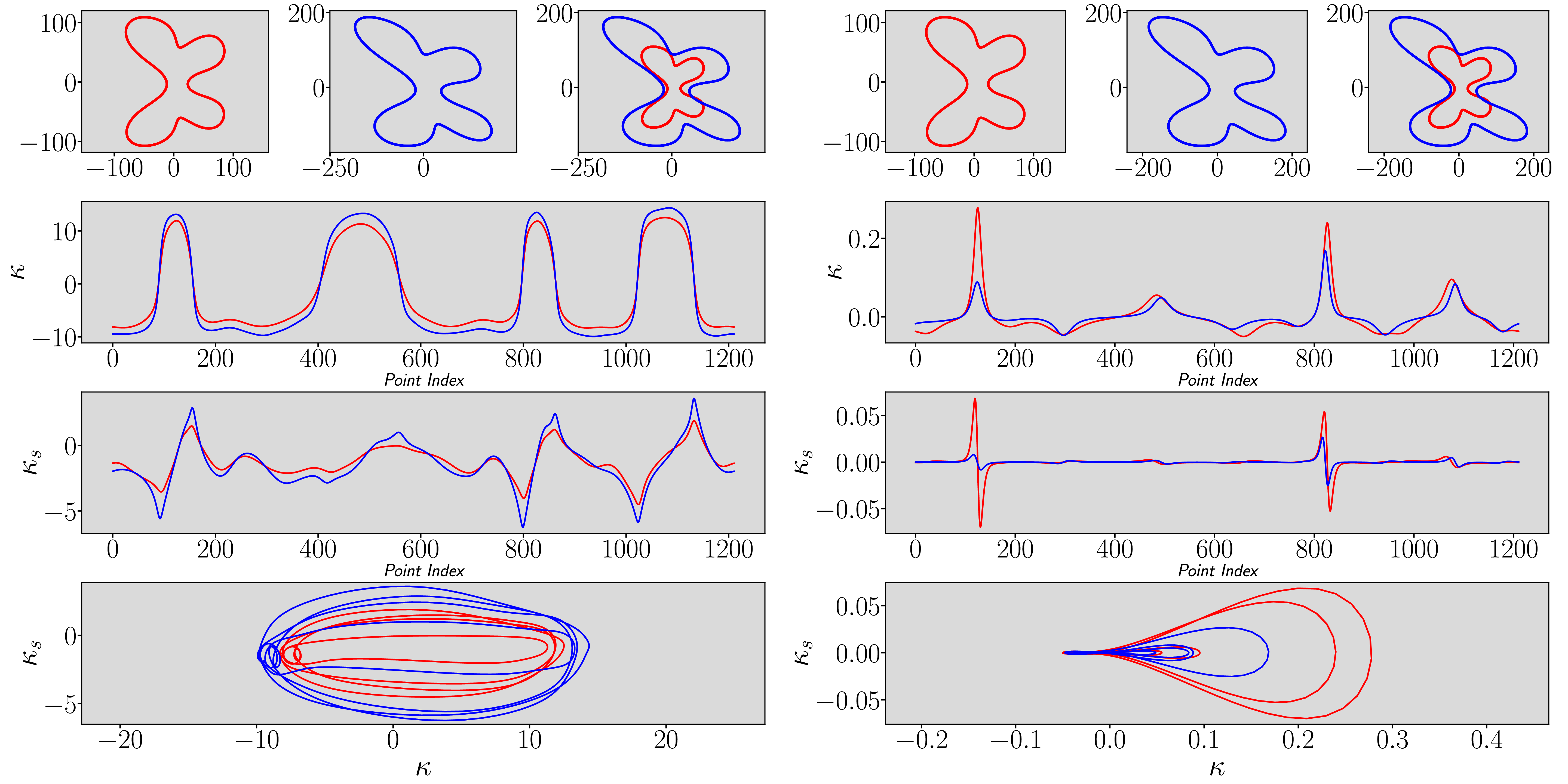}
                    \caption{Two affine equivalent curves (blue and red) and the approximation of their differential invariants. \textbf{Left}: The approximation of the affine $\kappa$ and $\kappa_s$ using our trained model. \textbf{Right}: The approximation of the Euclidean $\kappa$ and $\kappa_s$ using the axiomatic approach, which is based on finite differences.}
                    \label{fig:neural_vs_axiomatic}
                \end{figure*}

            \textit{Comparison with the Euclidean Axiomatic Approach.}
                As shown in Figure \ref{fig:neural_vs_axiomatic}, our model's approximations for $\kappa$ and $\kappa_s$ w.r.t. the affine group are superior over the axiomatic Euclidean approach (which is based on finite differences). As expected, the axiomatic signature exhibits visible misalignment of $\kappa$ and $\kappa_s$ w.r.t. the two equivalent curves, as well as major amplitude amplification differences. On the other hand, the approximations made by our model are aligned while only a mild amplitude amplification is visible.
        \subsection{Quantitative Evaluation}
\begin{figure*}[t!]
                \centering
                \includegraphics[width=\textwidth]{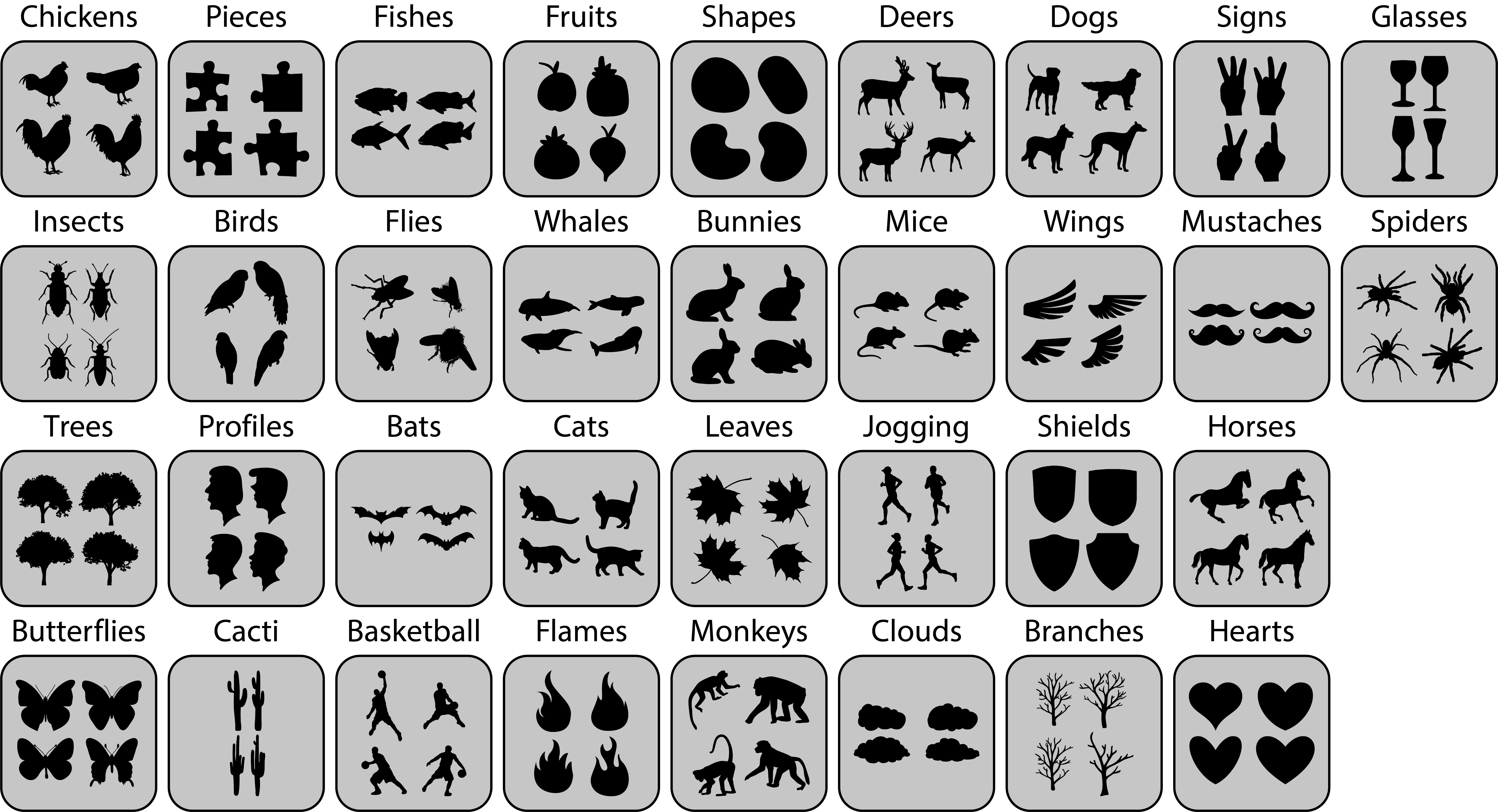} \\[0.5cm]
                \includegraphics[width=\textwidth]{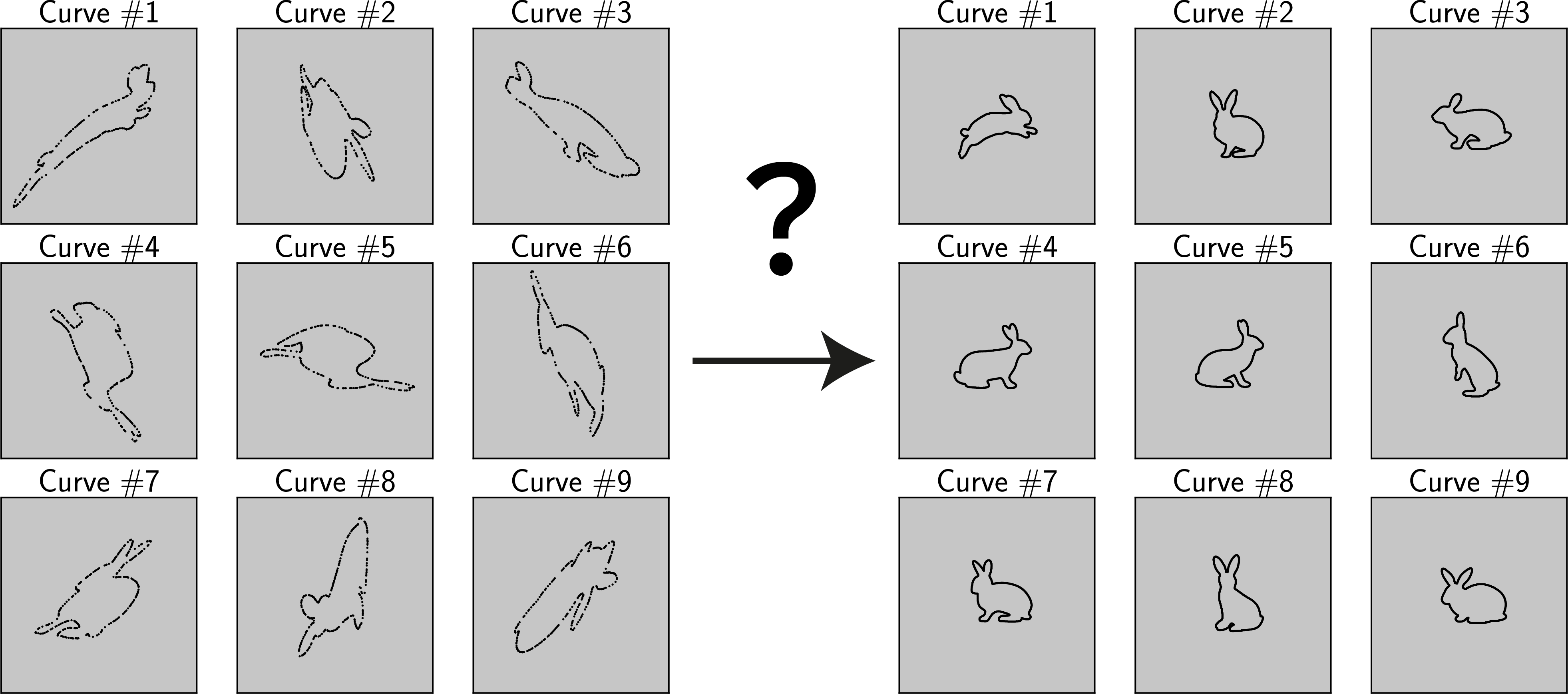}
                \caption{\textbf{Top}: A sample of our benchmark's curve collections. We plot four samples (out of 30) from each collection. \textbf{Bottom}: A demonstration of our shape-matching benchmark for the \textit{bunnies} collection. Each transformed and down-sampled curve on the left should be matched with its equivalent curve on the right. Although in this figure only 9 curves are shown, in the actual shape-matching benchmark each collection consists of 30 curves that need to be matched.}
                \label{fig:benchmark}
            \end{figure*}

            We have compiled an evaluation benchmark of 34 collections of planar curves, which were extracted from the boundary of silhouettes of natural objects, which are not necessarily smooth (see Figure \ref{fig:benchmark}). Each collection is comprised of 30 boundary curves of the same conceptual object in different poses and deformations.
            We perform a simple shape-matching evaluation which goes as follows. We iterate over all boundary curves in all collections. Given the $j^{\text{th}}$ curve in the $i^{\text{th}}$ collection, we deform it using an affine transformation and then down-sample it, in order to get what we call as a \textit{query} curve. Then, we compare the query curve against all curves in its collection (which we refer to as \textit{database} curves).
            A query curve is compared with a database curve by calculating the average Hausdorff distance of their corresponding signature curves, which were evaluated by our model. If the minimum distance was acquired by comparing a curve with its own transformed and down-sampled version, then we call it a \textit{successful match}. We score each collection by calculating the average number of successfully matched curves, which we denote by the \textit{success rate} of the collection.
            We perform this experiment under various down-sampling ratios and affine transformations.
            We use this benchmark to evaluate our model as well as the axiomatic approximations for the Euclidean and equiaffine groups, as proposed by \cite{Calabi98differentialand}.
            We run 3 different \textit{flavors} of our benchmark. For the first flavor, we transform each curve by an affine transform $A \in \mathbb{R}^{2 \times 2}$, where $\det\left(A\right) = 2$ and $\text{cond}\left(A\right) = 2$. For the second flavor we have $\det\left(A\right) = 2$ and $\text{cond}\left(A\right) = 3$, and for the third flavor we have $\det\left(A\right) = 3$ and $\text{cond}\left(A\right) = 2$. In all 3 flavors, we down-sample the query curves by increasing sampling ratios, where a sampling ratio of 100\% means that we keep all points, a sampling rate of 90\% means that we drop 10\% of the points, and so on. Evaluation results for the \textit{bunnies} collection are available in Table \ref{tbl:shape_matching}.
            \begin{table}[t!]
                \centering
                \caption{Shape-Matching Evaluation (Bunnies Collection)}
                \label{tbl:shape_matching}
                \resizebox{0.88\textwidth}{!}{%
                \begin{tabular}{|c|c|c|c|c|c|c|c|c|c|}
                \hline
                    \multicolumn{1}{|c|}{\textbf{}} & \multicolumn{9}{c|}{\textbf{Success Rate}} \\
                \hline
                    \multicolumn{1}{|c|}{} & \multicolumn{3}{c|}{\textbf{Our Model}} & \multicolumn{3}{c|}{\textbf{Euclidean Axiomatic}} & \multicolumn{3}{c|}{\textbf{Equiaffine Axiomatic}} \\
                \hline
                    \textbf{Sampling Rate} & \thead{$\det = 2$ \\ $\text{cond} = 2$} & \thead{$\det = 2$ \\ $\text{cond} = 3$} & \thead{$\det = 3$ \\ $\text{cond} = 2$} & \thead{$\det = 2$ \\ $\text{cond} = 2$} & \thead{$\det = 2$ \\ $\text{cond} = 3$} & \thead{$\det = 3$ \\ $\text{cond} = 2$} & \thead{$\det = 2$ \\ $\text{cond} = 2$} & \thead{$\det = 2$ \\ $\text{cond} = 3$} & \thead{$\det = 3$ \\ $\text{cond} = 2$} \\ \hline
                    100.00\% & 100.00\% & 100.00\% & 70.00\% & 3.33\% & 16.67\% & 10.00\% & 16.67\% & 10.00\% & 3.33\% \\ \hline
                    90.00\% & 100.00\% & 100.00\% & 63.33\% & 13.33\% & 6.67\% & 10.00\% & 3.33\% & 0.00\% & 3.33\% \\ \hline
                    80.00\% & 90.00\% & 90.00\% & 56.67\% & 10.00\% & 6.67\% & 10.00\% & 0.00\% & 0.00\% & 3.33\% \\ \hline
                    70.00\% & 83.33\% & 90.00\% & 46.67\% & 6.67\% & 0.00\% & 3.33\% & 3.33\% & 0.00\% & 0.00\% \\ \hline
                    60.00\% & 63.33\% & 73.33\% & 43.33\% & 13.33\% & 10.00\% & 6.67\% & 13.33\% & 3.33\% & 3.33\% \\ \hline
                    50.00\% & 56.67\% & 46.67\% & 33.33\% & 3.33\% & 3.33\% & 3.33\% & 3.33\% & 3.33\% & 3.33\% \\ \hline
                \end{tabular}}
            \end{table}
        \section{Conclusions and Future Research}
            We proposed a self-supervised approach, using multi-head Siamese neural networks, for the approximation of differential invariants of planar curves, with respect to a given transformation group $G$. 
            We have shown qualitatively that our model is invariant w.r.t. the group actions, as well as reparametrization and non-uniform down-sampling. We have also demonstrated the applicability of our method to the task of shape-matching.
            As a future research, we aim to design learning models for the approximation of higher dimensional differential invariants \cite{olver2009}, such as the torsion of space curves, and the principal curvatures of surfaces embedded in a Euclidean space.
\FloatBarrier
\bibliographystyle{splncs04}
\bibliography{mybibliography}
\end{document}